\documentclass{article}
\usepackage{spconf,amsmath,epsfig}
\usepackage{graphicx}
\usepackage{subfig}

\usepackage{smartdiagram}

\title{Integrating pre-processing pipelines in ODC based framework}
\twoauthors
    {U.Otamendi, I.Azpiroz, M.Quartulli, I.Olaizola}
        {Vicomtech Foundation \\ 
        \textit{Basque Research and Technology Alliance (BRTA)} \\ 
        Donostia-San Sebastián, 20009, Spain }{}{}
\begin{document}
%
\maketitle
\begin{abstract}

Using on-demand processing pipelines to generate virtual geospatial products is beneficial to optimizing resource management and decreasing processing requirements and data storage space. Additionally, pre-processed products improve data quality for data-driven analytical algorithms, such as machine learning or deep learning models. This paper proposes a method to integrate virtual products based on integrating open-source processing pipelines. In order to validate and evaluate the functioning of this approach, we have integrated it into a geo-imagery management framework based on Open Data Cube (ODC). To validate the methodology, we have performed three experiments developing on-demand processing pipelines using multi-sensor remote sensing data, for instance, Sentinel-1 and Sentinel-2. These pipelines are integrated using open-source processing frameworks.

\end{abstract}
\section{Introduction}
\label{sec:intro}

Geospatial imagery is widely used in multiple fields of environmental management approaches based on modern computing, such as deep learning \cite{zhu2017deep}. For instance, periodic data provided by satellites are useful for analysis and pattern extraction from a time series. The method offers a more accurate understanding of the evolution of the explored area. However, these high spatial resolution data require a large storage capacity. In addition, the processing of these data is computationally demanding \cite{qi2018board,thanh2018comparison}.

\begin{figure}[!ht]
\centering
    \includegraphics[width=0.9\linewidth]{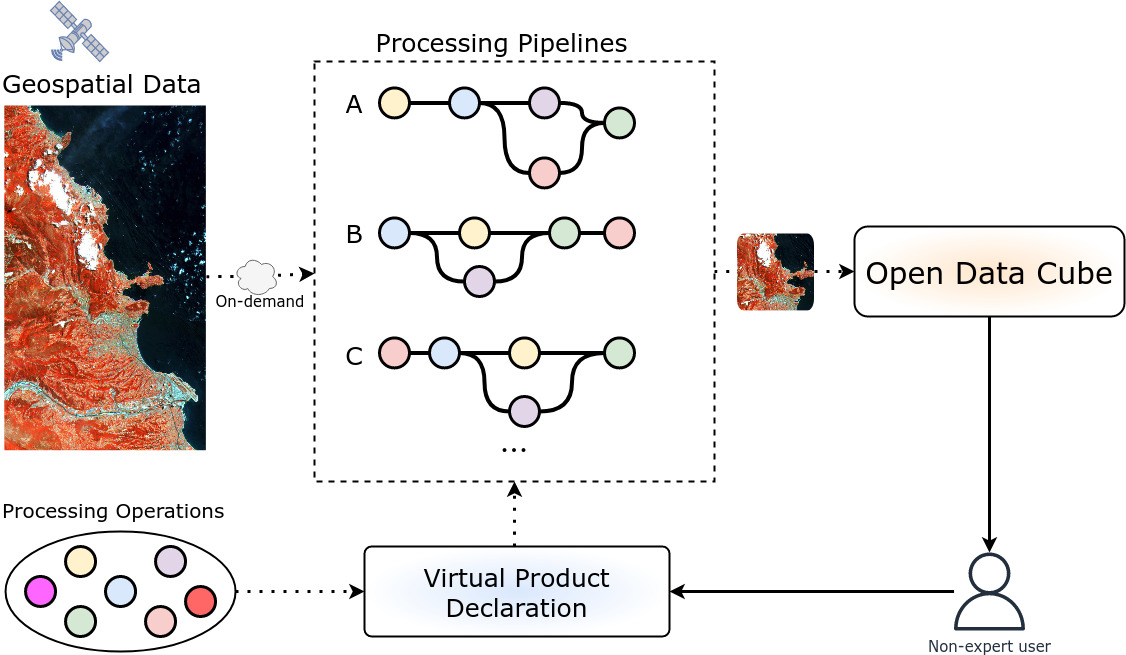}
\caption{The figure shows an overview of the proposed methodology to generate on-demand geospatial virtual products via processing pipelines. As shown, the non-expert user can declare a virtual product. Then, the framework uses the available processing operations to create a processing pipeline that converts the source geospatial data to the desired format. Finally, the resulting product is ingested by the Open Data Cube architecture, allowing the non-expert user to use the data in analytical processes.}
\label{fig:overview}
\end{figure}

Productive geo-imagery processing for rapid mapping is highly dependent on the efficiency of local statistics generation from remote sensing images. An automated computation supposes a substantial advance for agronomists, scientists, and satellite-derived data users.

In a previous paper \cite{otamendi2021geo} we proposed a methodology to address the limitations of non-expert users in managing and processing remote sensing and geo-imagery data. This system automatically ingests geospatial data and allows non-expert users to manage geospatial data in data-driven algorithms without requiring knowledge of remote sensing or geo-imagery exploitation. However, this considerably limits the exploration capability of modified products. Consequently, a non-expert user will only be limited to analyzing those products that the satellite imagery distributors have previously defined.

Therefore, this hinders the optimal use of the data in the performance of the algorithmic processes. In this sense, the main goal of the current contribution is to describe the integration of on-demand processing pipelines in an ODC-based infrastructure (see fig. \ref{fig:overview}). This approach provides several benefits of resource optimization and data quality improvement. Additionally, users acquire the ability to create \textit{virtual} geospatial data based on processing pipelines to automatically generate adequate data to train and use data-driven models \cite{ma2019deep}.

The implementation of this methodology has been integrated with the Open Data Cube (ODC) based architecture proposed in the previous data management paper \cite{otamendi2021geo}. In order to validate this approach, we have performed three experiments using different processing pipelines. \textbf{(1)} Cloud removal in Sentinel-2 imagery based on Weighted Average Synthesis Processor \textbf{(2)} Compute back-scattering in Sentinel-1 imagery based on the ESA Sentinel Applications Platform (SNAP) \textbf{(3)} Compute multiple remote sensing measurements indexes using Sentinel-2 imagery.

This paper is organized as follows:
Section \ref{sec:methodology} describes in detail the proposed methodology and integration. 
Section \ref{sec:experiment} presents details of the performed experiments and the validation. 
Finally, section \ref{sec:Conclusions} discusses results and includes the concluding remarks.

\section{Methodology}
\label{sec:methodology}

A \textit{virtual} product is a geospatial data that is generated with specific characteristics based on the source data, usually using raw data, based on a set of processing steps that provides the desired final data structure. The use of on-demand virtual products provides the capability to define a processing pipeline that will ingest geospatial data only when performing the generation. In terms of resource optimization, this will decrease data storage and management since the source data will only be downloaded when the processing of the desired area is requested. 

Additionally, using processing pipelines to generate \textit{virtual} products permits to automate of the processing step selection. Our methodology provides the ability to combine processing steps and evaluate the changes in the model behavior. Therefore, the proposed approach can be regarded as a data quality enhancement procedure.



In our previous work \cite{otamendi2021geo}, we proposed a methodology for geospatial data management and analysis, starting from product insertion to the loading of the raster in a georeferenced data frame. This approach supports the product metadata generation from multiple satellite data sources. In this paper, we have leveraged this aspect of the architecture, integrating the methodology mentioned above to provide the system with on-demand \textit{virtual} product management. The system that is used to automate the processing pipelines is based on scripts coded in the Python language that describe the steps involved and that can be coded with limited technical expertise.

This aspect of the integration provides added value to the data management and the produced data. This methodology considerably facilitates the operation of open-source processing algorithms and pipelines provided by the research community. In this sense, once the \textit{virtual} product generation is connected and correctly integrated into the processing pipeline, a non-expert user can produce the desired information without requiring knowledge about the underlying technology. 

Therefore, the approach proposed in this paper combines efficient resource optimization while allowing non-expert users to perform on-demand product generation based on open-source or manufactured processing pipelines.

\section{Experiment}
\label{sec:experiment}

This section briefly presents the details of the experiments performed to evaluate the proposed approach. To make a relevant analysis of the effectiveness of the framework, we have defined three realistic use cases: \textbf{1)} Cloud removal \textbf{2)} Compute back-scattering \textbf{3)} Compute multiple remote sensing measurements indexes. Additionally, to properly verify the viability of this approach, we have introduced the integration of third-party processing frameworks. This will allow non-expert users to take advantage of the processing algorithms created by the research community. The integration of these pipelines requires knowledge of ODC and the operation of the pipeline. Depending on the flexibility of each pipeline, the integration is more or less complex. Once the integration is completed, the operation does not require any expertise.

\subsection{Cloud Removal}

The task of cloud removal in geospatial imagery is widely used by the analytical pipelines \cite{zhang2019cloud,li2019cloud}. Usually, Sentinel-2 images contain significant cloud cover, hindering the correct analysis of the exploring areas. Therefore, cloud removal is beneficial for properly extracting patterns from multi-temporal remote sensing data. 

\begin{figure}[!ht]
\centering
    \subfloat[][Original image]{\includegraphics[width=0.415\linewidth]{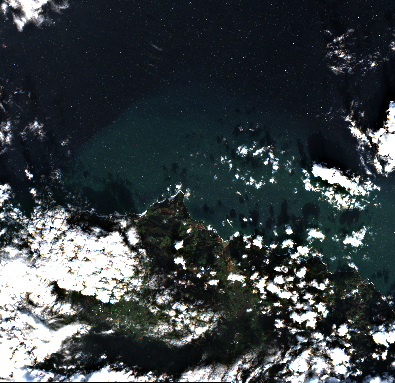}} \qquad
    \subfloat[][Processed image]{\includegraphics[width=0.45\linewidth]{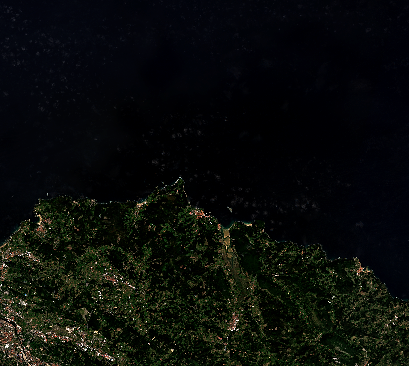}}
\caption{Illustration of the functioning of the integrated Weighted Average Synthesis Processor algorithm. On the left side an image of a original Sentinel-2 of the tile T30TWP. On the right side the cloud-free image created after the processing.}
\label{fig:wasp}
\end{figure}

In this case, we have integrated the open-source processing algorithm Weighted Average Synthesis Processor (WASP) provided by Theia. This Orfeo ToolBox (OTB) based processing chain creates monthly syntheses of cloud-free surface reflectances (see fig. \ref{fig:wasp}). 

In addition, this processing algorithm uses a specific data source generated by the MAJA processor based on Sentinel-2 data. In this sense, with this experiment, we have evaluated the behavior of the pipeline using third-party processing algorithms and \textit{non-standarized} data sources.

\subsection{Back-scattering analysis}
\label{sec:experiment2}

Back-scattering, also known as retro-reflection, is a physical phenomenon in which waves impacting a material at a certain angle are reflected at the same angle, back to the source from whence they originated. This phenomenon is usually explored in Synthetic-aperture radar (SAR) remote sensing data. The analysis of this aspect of the SAR imagery helps monitor different aspects of crop monitoring \cite{nasirzadehdizaji2021sentinel}, for instance, detecting flooded using classification models \cite{wang1995understanding}.

\begin{figure}[!ht]
\centering
    \includegraphics[width=1\linewidth]{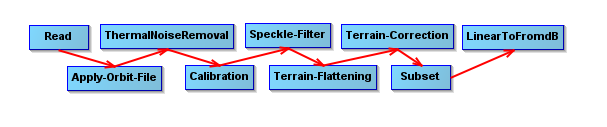}
\caption{Illustration of one of the back-scattering SNAP pipeline we have integrated in the system.}
\label{fig:snap1}
\end{figure}

In this experiment, to generate the product of back-scattering analysis, we integrated the ESA Sentinel Applications Platform (SNAP). This framework provides a product processing open toolbox for several satellites. In fact, we have used Snapista, which is a SNAP execution engine that facilitates the programmatic generation of SNAP GPT graphs (see fig. \ref{fig:snap1}). This engine supplies access to all functionalities derived from the toolboxes. Indeed, these graphs determine the necessary processing pipeline in order to obtain the expected product from distinct processed level satellite images (e.g., Sentinel 1, 2,3). 

\begin{figure}[!ht]
\centering
    \includegraphics[width=1\linewidth]{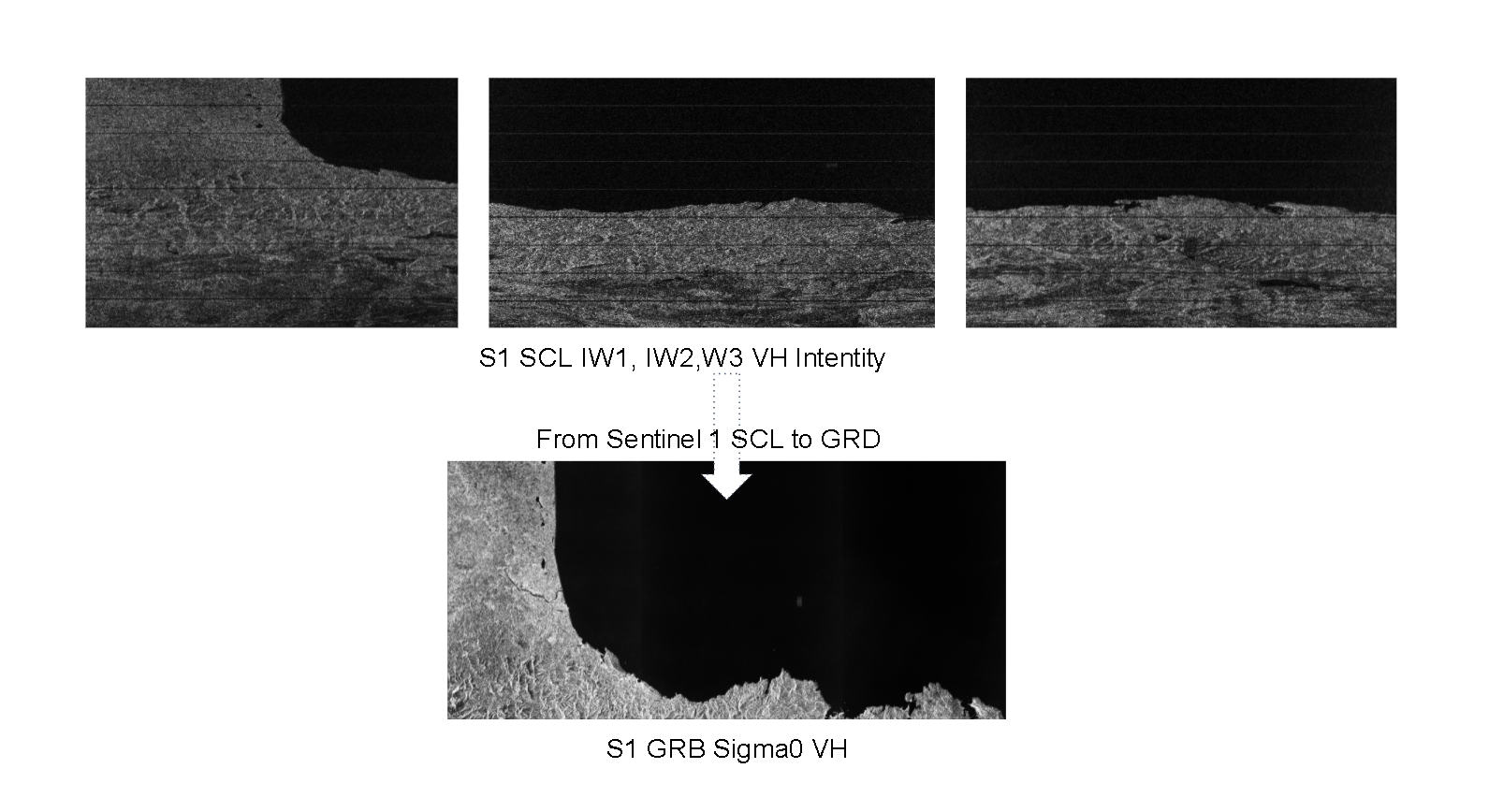}
\caption{Back-scattering coefficient analysis along the Biscay bay via Sentinel-1 imagery.}
\label{fig:snap}
\end{figure}

In this illustrative example, we consider the back-scattering procedure \ref{fig:snap1}. Once the data is requested, the required source data (Sentinel-1) is loaded on-demand. Straightaway, the Snapista engine loads and executes the processing pipeline defined in the declaration of the \textit{virtual} product. In this case, the processing is composed of different SNAP toolbox processing components, creating an adapted SNAP pipeline. The final result is visible in Figure \ref{fig:snap}.

This usage of the SNAP platform is extensible to several product-level processing pipelines. It facilitates the application of complex algorithms provided by the toolbox, which is helpful in remote sensing-based analysis.

\subsection{Compute measurements indexes}

The use of remote sensing data to detect changes in the ecosystem is a method that is being increasingly used in the literature \cite{pastick2019spatiotemporal,snyder2019comparison}. In this sense, measuring the spatiotemporal heterogeneity of ecosystem structure and function is a critical task to achieve. The novel method based on remote sensing permits the analysis of soil and plant indicators in vast regions in every part of the globe. This approach uses indexes computed from such geospatial data related to soil indicators of ecosystem health. 

\begin{figure}[!ht]
\centering
    \includegraphics[width=0.85\linewidth]{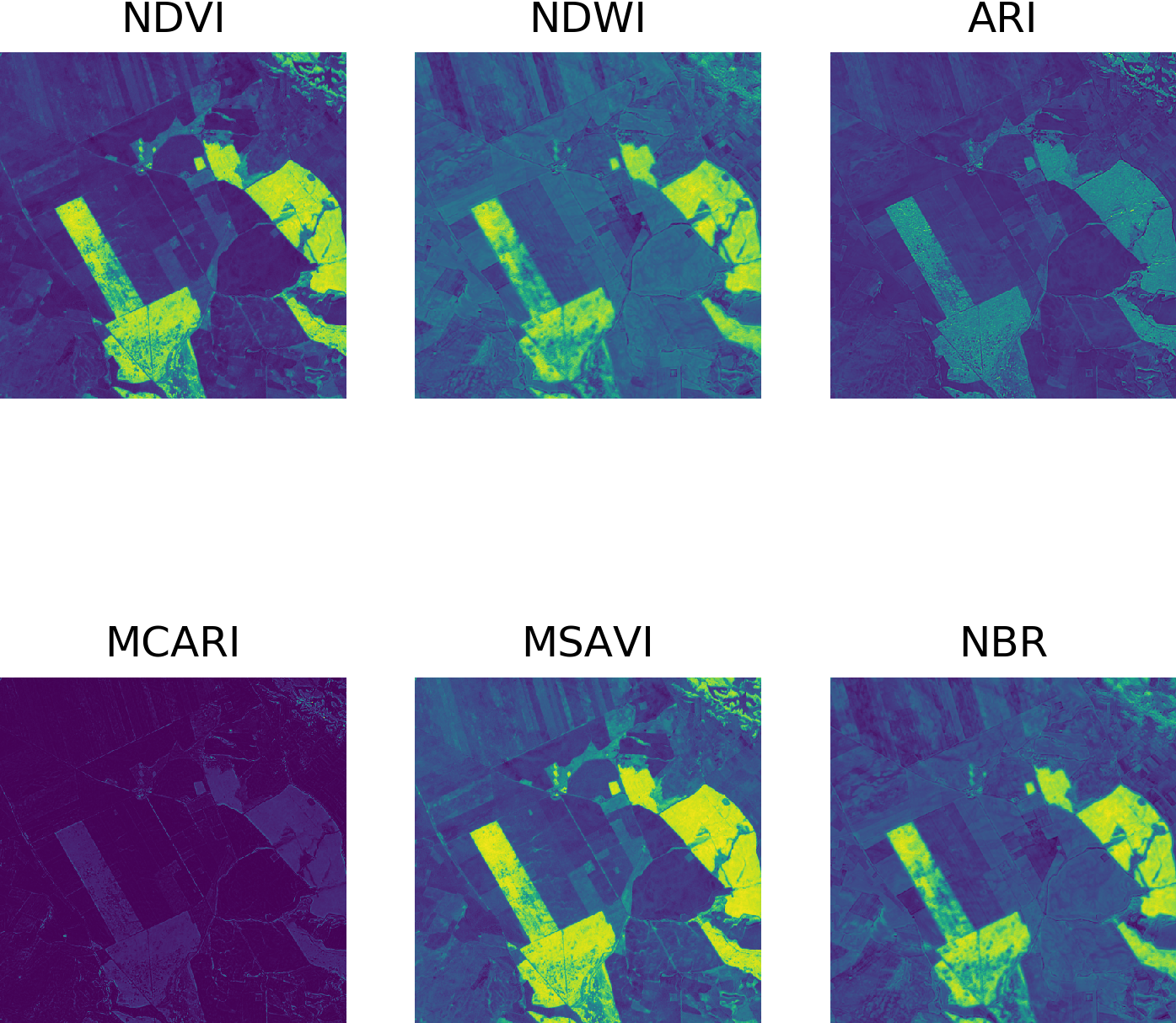}
\caption{Representation of some environment and ecosystem measurement indexes that are computed in the processing pipeline. The image shows a harvest field in the region of Aquitaine, France.}
\label{fig:indexes}
\end{figure}

In order to automatically generate the indexes used by the research community, we have experimented with integrating an index generation pipeline (see fig. \ref{fig:indexes}). The data source of this pipeline is the periodically captured Sentinel-2 data provided by ESA. This data is requested on-demand depending on the requisites of the virtual product defined by the user. 

As a matter of generalization, we have created a pipeline that generates the most widely used indexes: normalized difference vegetation index (NDVI), enhanced vegetation index (EVI), anthocyanin reflection index (ARI), modified soil-adjusted vegetation index (MSAVI), modified chlorophyll absorption in reflectance index (MCARI), structure insensitive pigment index (SIPI), normalized difference water index (NDWI) and normalized burn ratio (NBR).

\section{Conclusions}
\label{sec:Conclusions}

Current remote sensing data management strategies address the limitations of data storage and management, but not the ones related to data use and composition. In this paper, we provide an approach for a non-expert user to declare \textit{virtual} product based on processing pipelines. The approach allows exploring products that adjust to the analytical models' requirements and avoiding having to use only the products previously defined by remote sensing imagery distributors. 

To this end, we have conducted some experiments in which we have operated three on-demand processing pipelines integrated with open-source processing frameworks such as SNAP or WASP. The validation of this methodology has been performed in a geo-imagery management framework based in ODC \cite{otamendi2021geo}. The methodology has been integrated into an operational workflow of a center dedicated to the generation of geospatial products. The methodology has reduced the computational cost of product generation and the required storage capacity and facilitated process adaptability and monitoring.

In this context, the integration of processing libraries has incremented the capacity of the existent OCD-based satellite image ingestion service. Indeed, manually created virtual products from existing processed images can be extended to the complete procedure. Consecutive steps from data acquisition, image treatment procedures, data ingestion, computation of specific indicators (such as vegetation index), and other product creation are integrated into the presented system.

In addition, the implementation of pre-processing standard open-source libraries such as Snapista allows the option to define programmatically and execute user-adapted processing pipelines. This augments exponentially the capacity to manage different process level images and the derived results.

In future work, this methodology can be further extended to apply combinatorial exploration/optimization to generate processing pipelines. The procedure will lead to an automated generation and selection of the most adequate \textit{virtual} product in a cost-efficient manner. Additionally, using processing pipelines to generate \textit{virtual} products permits to automate of the processing step election. Artificial intelligence approaches can help identify the most appropriate combination of processing steps.



\bibliographystyle{IEEEbib}
\bibliography{strings,refs}

\end{document}